\documentclass{article}

\usepackage[dblblindworkshop, final]{neurips_2026}
\workshoptitle{Towards Test-Time Continual Learning Agents}

\usepackage[utf8]{inputenc} 
\usepackage[T1]{fontenc}    
\usepackage{hyperref}       
\usepackage{url}            
\usepackage{booktabs}       
\usepackage{amssymb}
\usepackage{amsfonts}       
\usepackage{nicefrac}       
\usepackage{microtype}      
\usepackage{xcolor}         
\usepackage{graphicx}
\usepackage{tabularx}
\usepackage{multirow}
\usepackage{tablefootnote}
\usepackage{enumitem}
\usepackage[most]{tcolorbox}
\usepackage{listings}

\title{Failure Broadens, Success Sharpens:\\Continual Adaptation of Tool-Use Agents Across Diverse Tasks with Dual-Polarity Memory}

\author{
  Tianmai M. Zhang$^{1}$\thanks{Work done during internship at Microsoft.} \qquad
  Hoda Shajari$^{2}$ \qquad
  Amr Hendy$^{2}$ \qquad
  Xiaohu Liu$^{2}$
  \\
  \textbf{Piyush Behre}$^{2}$ \qquad
  \textbf{Omar Zia Khan}$^{2}$ \qquad
  \textbf{Ankur Gupta}$^{2}$
  \\
  $^{1}$University of Washington \qquad
  $^{2}$Microsoft
  \\
  \texttt{tianmai@uw.edu} \qquad
  \texttt{hodashajari@microsoft.com}
}

\begin{document}
\maketitle

\begin{abstract}
Large language model agents can self-improve over time by utilizing procedural memory derived from prior task trajectories, yet the distinct utilities of memories derived from local successes and failures remain underexplored. This work studies this question in a realistic setting with heterogeneous streams of multi-domain tool-use tasks and various environment constraints common in real-world scenarios. We develop \emph{Dual-Polarity Memory with Periodic Maintenance} (DualPM), a simple, retrieval-free memory workflow characterized by (1) customizable reflection and procedural memory generation from both local successes and failures within each trajectory, and (2) periodic memory consolidation over long task streams under a given memory capacity, preventing memory explosion. Online prequential evaluation with fine-grained metrics reveals that failure memory more strongly broadens coverage, success memory better streamlines execution, while combining both polarities effectively integrates both benefits. Further studies demonstrate that DualPM enables cross-domain memory generalization and continual memory adaptation, while external evaluation on EvoMemBench reveals the portability and effectiveness of DualPM as a general long-term procedural memory method.
\end{abstract}

\section{Introduction}
\subsection{Background and related work}
As applications of large language model (LLM) agents increasingly emphasize utility in multi-session, long-horizon settings, externalizing experiences from past tasks as reusable memories or skills has become a common approach to enable continual personalization, adaptation, and improvement of AI agents~\citep{AgentMemorySurvey, ExperienceSpectrum}. Rather than retaining noisy raw interaction histories, training-free methods on agent memory or skills typically compress prior experience into knowledge or procedural guidance that can inform future task execution~\citep{AgentMemorySurvey}. 

A prominent strand of research in this direction derives reusable procedural guidance primarily from successful task executions. For example, Agent Workflow Memory~\citep{AWM} abstract successful trajectories into workflows; SkillWeaver~\citep{SkillWeaver} synthesizes tool interfaces from successful practice runs. Meanwhile, failed trajectories may carry distinct values that can be utilized through self-reflection. Reflexion~\citep{Reflexion}, an early pioneer in reflective self-learning, converts feedback from failed attempts into guidance for subsequent trails; recent work also showed that corrective updates derived from failures could improve computer-use agents~\citep{LearningFromFailures}. Recent trajectory learning approaches---including ExpeL~\citep{ExpeL}, ReasoningBank~\citep{ReasoningBank}, SkillOpt~\citep{SkillOpt}, MUSE~\citep{MUSE} and Trace2Skill~\citep{Trace2Skill}---draw on both successful and failed trajectories to derive reusable procedural guidance. These studies established that agents can self-improve at inference time by utilizing both previous successes and failures that are primarily judged and measured by benchmark-style, verifiable final task success.

Final-task success alone provides limited insight into whether memories derived from successful and unsuccessful behavior serve different functions or how they affect actions within a multi-step trajectory. Successes and failures can also occur locally within a trajectory rather than only at the episode level \citep{ReMe}. A successful trajectory may contain local failures worth avoiding or improving, while a failed trajectory may contain effective behaviors worth reusing. Furthermore, the explicit outcome signals available in benchmark environments are often absent in the real world. In real-world user-facing scenarios, agents may also face incomplete environment information, noisy tool interfaces, ambiguous user requests, and domain-diverse task streams. Under these conditions, deriving and maintaining a durable pool of procedural guidance over long, heterogeneous task streams remains challenging and underexplored.

\subsection{Objectives and Contributions}
This work studies cross-session procedural memory in a realistic setting where an agent handles a heterogeneous stream of multi-step, multi-domain tool-use tasks under various constraints: user requests may be implicit; the tool set is large, noisy, and not fully visible upfront; evaluation ground truth is not available to the memory workflow; and the memory pool has a fixed capacity. The memory mechanism must therefore determine procedural lessons that remain durable and applicable to tasks from different domains, and continuously maintain the memory pool based on observations from newly completed task episodes.

Our primary research question is how success and failure memories, independently and jointly, affect agent performance. To answer this, we introduce \emph{Dual-Polarity Memory with Periodic Maintenance} (DualPM), a simple, retrieval-free memory workflow that reflects on completed task episodes, drafts success and failure memories, and periodically consolidates them into separate memory pools. Through online, prequential evaluation and fine-grained measurements of intermediate task stages, we find that the two memory polarities have complementary utilities: failure memory broadens coverage more, success memory improves efficiency more, and their combination effectively integrates both benefits.

Case studies and analyses demonstrate that DualPM supports continuous memory adaptation after the procedural memory pool of a user is transferred to another user task sequence with a distinct domain composition. External evaluation on EvoMemBench \citep{EvoMemBench} further reveals the effectiveness of DualPM as a portable long-term procedural memory method that works across different task environments and backbone models.




\section{Methodology}
\subsection{The DualPM workflow}
\label{sec:DualPM}

\paragraph{Problem formulation} We consider memory-driven cross-session continual learning over an ordered task sequence $\{t_i\}_{i=1}^n=(t_1,t_2,\ldots,t_n)$. For each task $t_i$, an agent produces an observed trajectory $\tau_i\sim\pi(t_i,\mathcal{M}_i)$, conditioned on an active memory context $\mathcal{M}_i$. The objective of a test-time memory workflow is to transform earlier trajectories $\{\tau_j\}_{j<i}$ into a memory context $\mathcal{M}_i$ that improves the execution of subsequent tasks $\{t_j\}_{j\geq i}$, without updating the agent model, harness, or environment settings. \emph{Dual-Polarity Memory with Periodic Maintenance} (DualPM) operationalizes this process through periodic memory generation and parallel maintenance of success memory $\mathcal{M}^+$ and failure memory $\mathcal{M}^-$.

\begin{figure}[h]
  \centering
  \includegraphics[width=\textwidth]{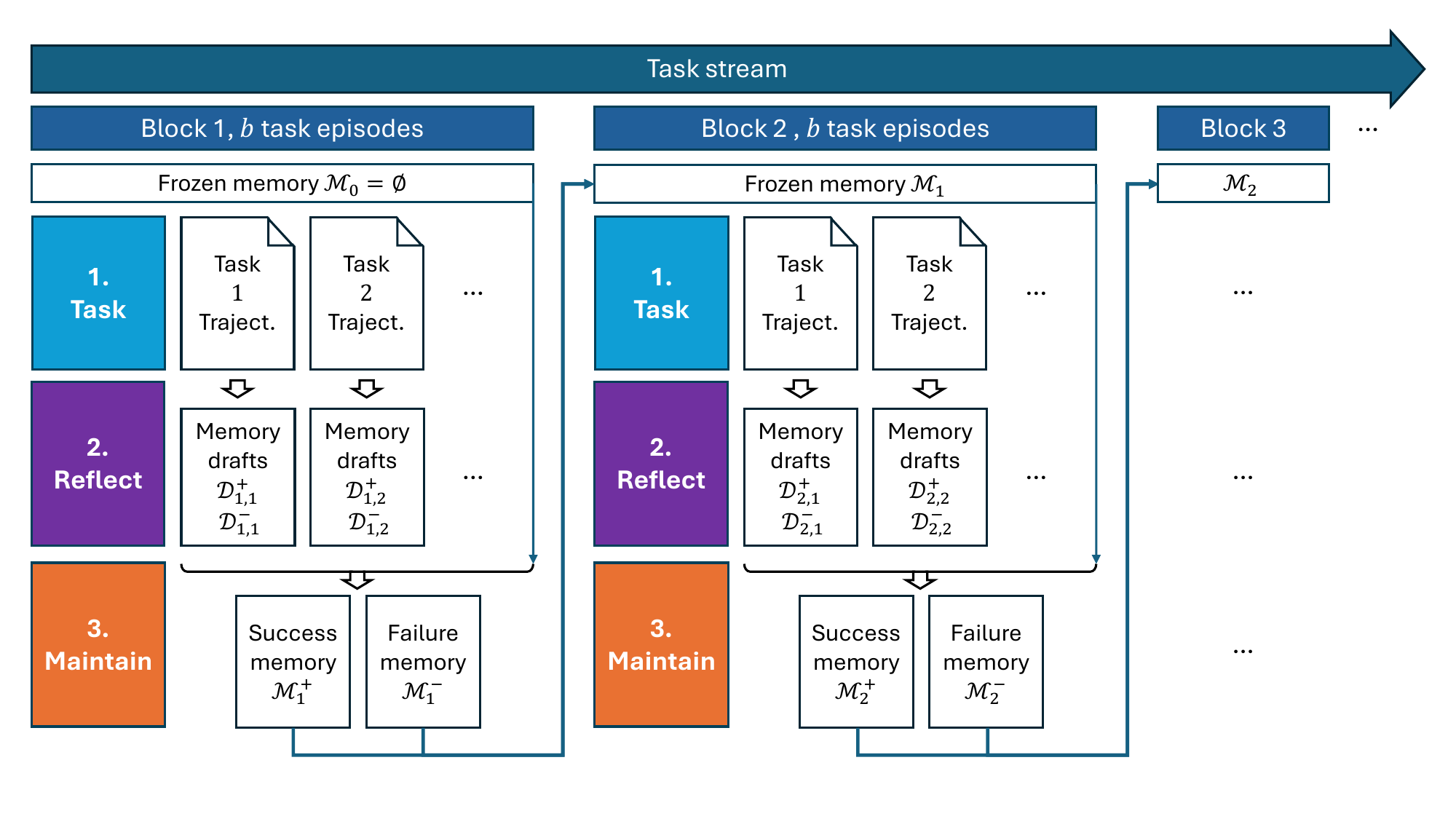}
  \caption{Blockwise task--reflection--maintenance workflow of DualPM.}
  \label{fig:DualPM}
\end{figure}

\paragraph{Blockwise task--reflection--maintenance cycle}
DualPM partitions the task sequence $\{t_i\}_{i=1}^n$ into consecutive blocks of size $b$, $\mathcal{B}_k=\{t_{(k-1)b+1},\ldots,t_{\min(kb,n)}\}$ for $k=1,2,\ldots,\lceil n/b\rceil$. Each block follows three sequential phases (Figure~\ref{fig:DualPM}). During the \emph{task phase}, every task in $\mathcal{B}_k$ is executed with the same frozen memory context $\mathcal{M}_{k-1}$, producing trajectories $\{\tau_i:t_i\in\mathcal{B}_k\}$. During the \emph{reflection phase}, DualPM employs an LLM to first reflect on each completed trajectory, and then generate drafts of success and failure memories, denoted as $\mathcal{D}_k^{+}$ and $\mathcal{D}_k^{-}$. During the \emph{maintenance phase}, these memory drafts are consolidated independently into success and failure memory lists, producing $\mathcal{M}_k^+$ and $\mathcal{M}_k^-$. The aggregated memory list $\mathcal{M}_k=\mathcal{M}_k^+\bigcup\mathcal{M}_k^-$ becomes the memory context for tasks in the next block $\mathcal{B}_{k+1}$. The initial block $\mathcal{B}_1$ cold-starts with no memory $\mathcal{M}_0=\varnothing$.

\paragraph{Reflection and memory drafting}
During the reflection phase, DualPM makes two sequential LLM calls for each completed trajectory $\tau_i$ associated with task $t_i\in\mathcal{B}_k$. The first call performs reflection on the observed trajectory $\tau_i$, including the agent-visible instructions, user request, agent actions, tool responses, and final agent answer. Task ground truth, such as the golden trajectory and reference answer, is never exposed to the reflection model. The reflection prompt includes a customizable feedback signal section specifying what constitutes granular success or failure within the trajectory, thereby guiding the reflection process.

The second LLM call continues from the reflection and drafts reusable procedural lessons for trajectory $\tau_i$. It may generate lessons of either polarity: success memory drafts $\mathcal{D}_{k,i}^{+}$ capture reusable principles for effective behaviors worth repeating, whereas failure memory drafts $\mathcal{D}_{k,i}^{-}$ provide constructive guidance for avoiding or improving an ineffective behavior. The memory drafting call produces up to $d$ drafts for each polarity. All memory drafts adopt the following structure:
\begin{itemize}[leftmargin=*,nosep, topsep=1pt]
    \item \texttt{memory\_type}: success memory or failure memory;
    \item \texttt{title}: a short, transferable name for the procedural principle;
    \item \texttt{content}: actionable guidance for future tasks and its rationale;
    \item \texttt{evidence}: a concise account of the trajectory's evidence that supports the guidance;
    \item \texttt{applicability}: an access-time routing cue specifying when the guidance should be considered;
    \item \texttt{confidence}: qualitative assessment of the drafting model's confidence in deriving the guidance.
\end{itemize}
Each draft additionally records a draft index and source-episode reference for provenance. All prompts are available in Appendix~\ref{app:DualPM_prompts}.

\paragraph{Memory maintenance}
After reflection on block $\mathcal{B}_k$, DualPM consolidates the resulting drafts while maintaining the two memory polarities independently. For each polarity $p\in\{+,-\}$, a separate LLM call receives the previous memory list $\mathcal{M}_{k-1}^{p}$ together with the new drafts $\mathcal{D}_k^{p}$, then returns the complete updated list $\mathcal{M}_k^{p}$, subject to a capacity constraint $|\mathcal{M}_k^{p}|\leq m$. This constraint encourages draft consolidation and evidence-based revision: rather than simply appending new memories, the maintainer must keep the memory list concise and nonredundant by preserving, revising, merging, splitting, creating, or removing memories based on new drafts and evidence from the latest block. Each maintained memory unit follows the draft schema above but may retain up to $e$ supporting evidence items selected from its parent memories or newly incorporated drafts. Index references to these sources are recorded for provenance. The updated memory lists are then combined as $\mathcal{M}_k=\mathcal{M}_k^{+}\bigcup\mathcal{M}_k^{-}$ and applied to the next block $\mathcal{B}_{k+1}$.

\subsection{The TRAJECT-Bench environment and agent runtime}
\label{sec:Traject_env}
To evaluate agent performance and task progress in a granular manner, we use TRAJECT-Bench \citep{TrajectBench}, a trajectory-aware tool-use benchmark providing diverse user requests spanning multiple domains and ground-truth tool trajectories for the tasks. Each user request in TRAJECT-Bench consists of multiple subtasks, typically requiring 3 to 10+ different tool calls to the gold tools \citep{TrajectBench}. To complete the user request, an agent needs to select from a pool of hundreds of candidate tool APIs provided for the task's domain, and make function calls to those tools using correct parameter combinations inferred from the user request \citep{TrajectBench}. The large candidate tool pool also simulates the real-world scenario where an agent is exposed to numerous tools provided by Model Context Protocol servers. Some task examples in TRAJECT-Bench can be found in Appendix~\ref{app:Traject_examples}.

TRAJECT-Bench primarily evaluates LLMs through one-shot prediction, with all tool schemas supplied in the prompt. Since this does not reflect the task environment faced by modern agents, we construct a more realistic, single-agent agent runtime with provider-native tool provision, parallel tool calling, and ReAct-style \citep{ReAct} agent-tool interaction. During task execution, a tool would replay its gold response recorded in the ground-truth dataset if the agent calls it using the correct parameter combination for required parameters. Tool calls to either non-gold tools or gold tools but with incorrect parameters, however, would receive a general failure message that instructs the agent to reconsider its tool or parameter selection. This replay-based procedure would naturally result in multiple trials and errors until the agent makes a correct tool call, while ensuring that final agent responses in our experiments are comparable to the reference answers provided by TRAJECT-Bench.

We further increase task difficulty through two design choices motivated by real-world agent applications. First, we use only the hard version of TRAJECT-Bench, which is designed to mimic real-world interactions where users often describe their requests colloquially and implicitly \citep{TrajectBench}. Second, modern agent applications (e.g., ChatGPT, Codex, and M365 Copilot) commonly do not expose all available tools to the agent upfront, but instead provide only a short list of core built-in tools plus a tool discovery function that retrieves additional tools when necessary. To mimic this design, we hide each task's candidate tool pool behind a \texttt{tool\_search} function that the agent can use to retrieve tools relevant to a user request. The function ranks candidate tools through BM25 matching between the agent's search query and tool metadata, and then returns the requested number of top-ranked tool schemas. Retrieved tools are added to the agent’s callable tool set and remain available throughout the episode. The agent may invoke \texttt{tool\_search} repeatedly to discover additional tools, but cannot call a tool before retrieving it. The system prompt for the task agent is available in Appendix~\ref{app:Traject_prompts}.

\subsection{Stage-aware evaluation}
\label{sec:eval_metrics}
To effectively solve a user request in the revised TRAJECT-Bench environment, an agent needs to complete three intermediate stages of execution. First, it must \emph{discover} the appropriate tools by issuing effective \texttt{tool\_search} queries. This is challenging because implicit user requests often describe only the required functionality (e.g., verifying an email address) rather than naming the tool directly (e.g., \texttt{Blaze Verify} in Appendix~\ref{app:Traject_examples}, task example 1). Second, it must \emph{select} tools from the retrieved schemas that correspond to the request’s sub-requirements. Third, it must \emph{use} each tool correctly by supplying ground-truth-matching parameters to obtain a valid response, which then serves as evidence for the final answer. For each task episode, we evaluate this three-stage funnel using the following metrics:

\begin{itemize}[leftmargin=*,itemsep=0.5pt,topsep=1pt]
    \item \textit{Discovery recall}: The proportion of ground-truth tool APIs discovered through \texttt{tool\_search}. A higher discovery recall indicates better discovery coverage.
    \item \textit{Selection recall}: The proportion of ground-truth tool APIs invoked by the agent. A higher selection recall indicates better selection coverage. This metric is identical to TRAJECT-Bench's \textit{Inclusion}. 
    \item \textit{Usage recall}: The proportion of ground-truth tool calls matched by tool calls invoked by the agent. A higher usage recall indicates better tool execution coverage and overall task progress. This metric is revised from TRAJECT-Bench's $usage$, which is conditional on selecting a ground-truth tool.
    \item \textit{Usage precision}: The proportion of the agent's task-relevant tool calls that match ground-truth calls. Any incorrect, malformed, or redundant calls would reduce precision.
    \item \textit{Total turns}: The number of agent interaction turns until it produces the final answer. Fewer turns indicate higher trajectory efficiency. This metric is also a proxy for task latency, as each turn corresponds to an LLM call. 
    \item \textit{Final answer accuracy}: Binary final answer correctness judged by an LLM against the reference answer in TRAJECT-Bench, complementary to the deterministic \textit{usage recall}. This metric is inherited from TRAJECT-Bench. Only answers successfully addressing all subtasks in the user requests would be considered correct; partially correct answers are considered incorrect.
\end{itemize}

\subsection{Profile-based task sequence construction}
\label{sec:data_construction}

To simulate task diversity in the real world, we construct four mixed-domain task sequences by reorganizing the domain-specific tasks in the parallel-hard subset of TRAJECT-Bench. Each task sequence contains 200 unique tasks sampled from two primary domains and three secondary domains based on a synthetic user profile (Table~\ref{tab:user_domain_setup}). Matching DualPM's blockwise design, each task sequence is organized into ten 20-task blocks with a consistent domain distribution, i.e., 70\% from primary domains and 30\% from secondary domains. Task sequences are frozen for the experiments.

\begin{table*}[h]
\centering
\caption{User profiles and domain composition of the four task sequences. Primary domains constitute 70\% of a user's task sequence, while secondary domains constitute 30\%.}
\label{tab:user_domain_setup}
\setlength{\tabcolsep}{8pt}
\begin{tabular}{llll}
\toprule
User & Profile & Primary domains & Secondary domains \\
\midrule
U1 & Information worker
& Email, News/Media
& Finance, Education, Weather \\
U2 & Student
& Education, Email
& Gaming, E-commerce, Music \\
U3 & Budget-aware shopper
& E-commerce, Finance
& Travel, Weather, Education \\
U4 & Entertainment lover
& Gaming, Music
& E-commerce, Travel, Finance \\
\bottomrule
\end{tabular}
\end{table*}

\section{Experiments and results}
\subsection{Memory polarity experiment}
\label{sec:memory_polarity}
The main experiment examines the independent and joint effects of success- and failure-derived procedural memory. We adopt a $2\times2$ factorial design involving four memory conditions: no memory, failure memory only, success memory only, and both on. During the task phase, the agent receives the complete, active memory list permitted by its assigned condition, appended to the system or developer prompt as memory context (see prompt template in Appendix~\ref{app:Traject_prompts}). We use GPT-5.5 as the backbone model for both the task agent and the DualPM memory workflow, and DeepSeek-V3.2 as the judge for final answer accuracy to avoid evaluation with the same model. Details of model setup are available in Appendix~\ref{app:supplemental_results}.

The experiment evaluates task agent performance in an online, prequential manner. Under DualPM's periodic memory update workflow (Section~\ref{sec:DualPM}), tasks in each block are first used for testing and then included in training as they enter memory generation and maintenance. With $n=200$ tasks in each user's task sequence, we set block size $b=20$ for DualPM, yielding a total of ten blocks and an evaluation sample size of nine blocks, $180$ tasks. Across four user task sequences, this yields 720 evaluated task episodes per condition. We heuristically set memory capacity parameters to $d=2$, $m=10$, and $e=3$, i.e., each episode can produce up to 2 drafts per polarity; the maintained memory list contains up to 10 memory units per polarity, with each memory unit containing up to 3 evidence items. This parameter setting effectively caps the memory list generated by GPT-5.5 at around five thousand tokens per polarity. For the same user sequence, all three memory-on conditions reuse the same block 1 from the no-memory run to ensure a fair cold start, and then fork with their own memory exposure settings. Statistical comparisons of the memory conditions are based on paired differences by task within each user sequence. Details of statistical methods are available in Appendix~\ref{app:supplemental_results}.

\begin{table*}[h]
\centering
\caption{Performance of memory polarities averaged across all four users. Recalls, usage precision, and final answer accuracy are reported in percentages. \textbf{Bold} and \underline{underline} mark the best and second-best values, respectively.}
\label{tab:memory_polarity_results}
\setlength{\tabcolsep}{6pt}
\begin{tabular}{lcccccc}
\toprule
\multirow{2}{*}[-0.6ex]{\shortstack[l]{Memory\\Setting}}
& \multicolumn{3}{c}{Recall}
& \multicolumn{2}{c}{Efficiency}
& \multirow{2}{*}[-0.6ex]{\shortstack[c]{Answer\\Accuracy}} \\
\cmidrule(lr){2-4}
\cmidrule(lr){5-6}
& Discovery
& Selection
& Usage
& Usage Precision
& Turns
&  \\
\midrule
No memory
& 75.2 & 68.0 & 66.0 & 50.5 & 6.1 & 68.1 \\
Failure only
& \underline{81.0} & \underline{73.9} & \underline{71.9}
& 49.0 & 6.3 & 71.0 \\
Success only
& 78.7 & 72.2 & 70.5
& \textbf{56.2} & \textbf{5.4} & \underline{72.2} \\
Both on
& \textbf{81.1} & \textbf{74.5} & \textbf{72.4}
& \underline{53.7} & \underline{5.5} & \textbf{73.2} \\
\bottomrule
\end{tabular}
\end{table*}

\textbf{Failure memory primarily broadens coverage, success memory primarily streamlines execution, and their combination integrates both benefits.} Table~\ref{tab:memory_polarity_results} shows that all three memory-on settings improved tool utilization coverage and final answer accuracy over the no memory setting. The improvements are statistically significant across all three recall measures (all $p\leq .001$) and final answer accuracy ($p\leq .015$) in paired comparisons across all four users. The two memory polarities exhibit distinct utility profiles. Failure-only memory achieved significantly higher discovery and selection recall than success-only ($p=.002$ and $p=.018$, respectively). This pattern indicates that failure lessons primarily broaden the set of relevant capabilities considered and selected by the agent. Success-only memory instead produced the highest tool usage precision and completed the tasks in fewest turns (both $p<.001$ compared to failure-only).

Combining both polarities integrated these complementary effects and achieved the best performance overall. Relative to success-only memory, the combined condition significantly improves discovery, selection, and usage recalls (all $p\leq .002$) with only slightly lower precision ($p<.001$) and comparable total turns ($p=.27$). Relative to failure-only memory, it achieves comparable or modestly better recalls ($p=.41$, $.038$, and $.045$ for discovery, selection, and usage recalls, respectively) while significantly improved usage precision ($p<.001$) and task completion in fewer turns ($p<.001$). In terms of final answer accuracy, the both-on condition achieved the highest score, although the differences among the memory-on settings were not statistically significant (all $p\geq.17$). Together, dual-polarity memory achieved the strongest overall coverage while retaining comparable tool usage precision, interaction length, and answer accuracy.

\textbf{Memory promoted more targeted tool searches and more judicious tool calls.} Paired trajectory analysis indicates that memory changed how the agent organized tool discovery and execution. Memory-on trajectories more often decomposed multi-part requests into subtask-specific \texttt{tool\_search} queries rather than relying on a single generic query, resulting in more tool search actions and lower (better) mean ranks of gold tools in tool search results (Appendix~\ref{app:supplemental_results}, Table~\ref{tab:memory_polarity_results_sup}). For example, on a composite, class-presentation preparation task (task example 2 in Appendix~\ref{app:Traject_examples}) requesting a fact about number 42, DEI (diversity, equity, and inclusion) learning resources, and a usage example for the word ``excellent'', the no-memory agent issued one broad query: ``web search internet search retrieve links snippets definitions dictionary'', which retrieved a dictionary tool but no tools related to the other two subtasks. Then, the no-memory agent made no further searches, answered from prior knowledge, and invented plausible links to DEI resources. Instead, the agent with dual-polarity memory issued separate tool search queries for ``Numbers trivia fact number 42'', ``DEI learning resources page links'', and ``dictionary definition word excellent usage'', allowing it to obtain tools for all three subtasks. The memory lesson that potentially drove this improvement can be found in Appendix~\ref{app:memory_examples}.

The paired trajectories further illustrate the distinct utilities of the two memory polarities. In a task requiring ten pieces of evidence across multiple topics, the no-memory agent issued one generic search and selected tools covering only 20\% of the request requirements. With failure-only memory, active memory guidance emphasized decomposing the request by evidence type and recovering unresolved material requirements. The agent consequently pursued five subtask-oriented searches and expanded execution to 16 tool calls over ten turns, increasing selection recall to 80\%. Success-only memory achieved the same 80\% selection recall with 16 calls but only five turns, and a usage precision of 80\%. Although no trajectory produced a fully correct final answer for the task, this contrast demonstrates the broader pattern reflected in Table~\ref{tab:memory_polarity_results}: failure memories operated primarily as coverage-and-recovery checklists that encouraged continued pursuit of incomplete subtasks, whereas success memories operated more as routing guidance that organized parallel work into a more compact execution trajectory. Similar to the success-memory agent, the both-on agent completed the task in five turns but with a 100\% precision (i.e., no failed tool call). Combining performance results with token usage metrics (Appendix~\ref{app:supplemental_results}, Table~\ref{tab:memory_polarity_results_sup}) reveals a practical performance-cost trade-off: success-only is preferable when prioritizing execution efficiency and cost, whereas dual-polarity memory is preferable when maximizing coverage and performance.

\subsection{Case studies}
\label{sec:case_studies}
Following the main experiment, we conducted case studies to further examine cross-domain memory transfer and memory evolution patterns.

\paragraph{Case study 1: cross-domain memory transfer} To examine whether procedural memories learned from one user remain useful on a largely distinct task sequence, we select U1 and U4, as this pair has minimal overlap in task domains and is therefore the most challenging for cross-domain evaluation. Memory transfer is evaluated in both directions (U1 $\to$ U4 and U4 $\to$ U1). For each target sequence, we compare four settings: (1) no memory; (2) standard DualPM memory from scratch; (3) the other user's final dual-polarity memory applied to all target tasks but kept frozen; and (4) the same transferred memory used as a warm start with standard DualPM updates afterwards. Results in Table~\ref{tab:cross_user_results} shows that frozen procedural memory learned from different domains yielded moderate performance gains, with improvements in all recall measures and final accuracy. Enabling memory adaptation led to continual update of the transferred memory and yielded comparable performance to building memory from scratch.

\begin{table*}[h]
\centering
\caption{Cross-user memory transfer effects, averaged across U1 $\to$ U4 and U4 $\to$ U1. Recalls, usage precision, and final answer accuracy are reported in percentages. \textbf{Bold} and \underline{underline} mark the best and second-best displayed values of memory-on settings, respectively; ties are marked equally.}
\label{tab:cross_user_results}
\setlength{\tabcolsep}{6pt}
\begin{tabular}{lcccccc}
\toprule
\multirow{2}{*}[-0.6ex]{\shortstack[l]{Memory\\Setting}}
& \multicolumn{3}{c}{Recall}
& \multicolumn{2}{c}{Efficiency}
& \multirow{2}{*}[-0.6ex]{\shortstack[c]{Answer\\Accuracy}} \\
\cmidrule(lr){2-4}
\cmidrule(lr){5-6}
& Discovery & Selection & Usage & Usage Precision & Turns & \\
\midrule
No memory
& 79.0 & 71.9 & 69.6
& 49.1 & 6.2 & 72.0 \\
Own memory
& \textbf{82.1} & \textbf{76.2} & \textbf{73.8}
& \textbf{50.6} & \textbf{5.6} & \underline{78.0} \\
Transferred, frozen
& 79.3 & 73.7 & 71.5
& 48.2 & 6.4 & 75.0 \\
Transferred, adaptive
& \textbf{82.1} & \underline{75.3} & \underline{73.5}
& \underline{50.0} & \textbf{5.6} & \textbf{78.5} \\
\bottomrule
\end{tabular}
\end{table*}

\begin{figure}[h]
  \centering
  \includegraphics[width=0.95\textwidth]{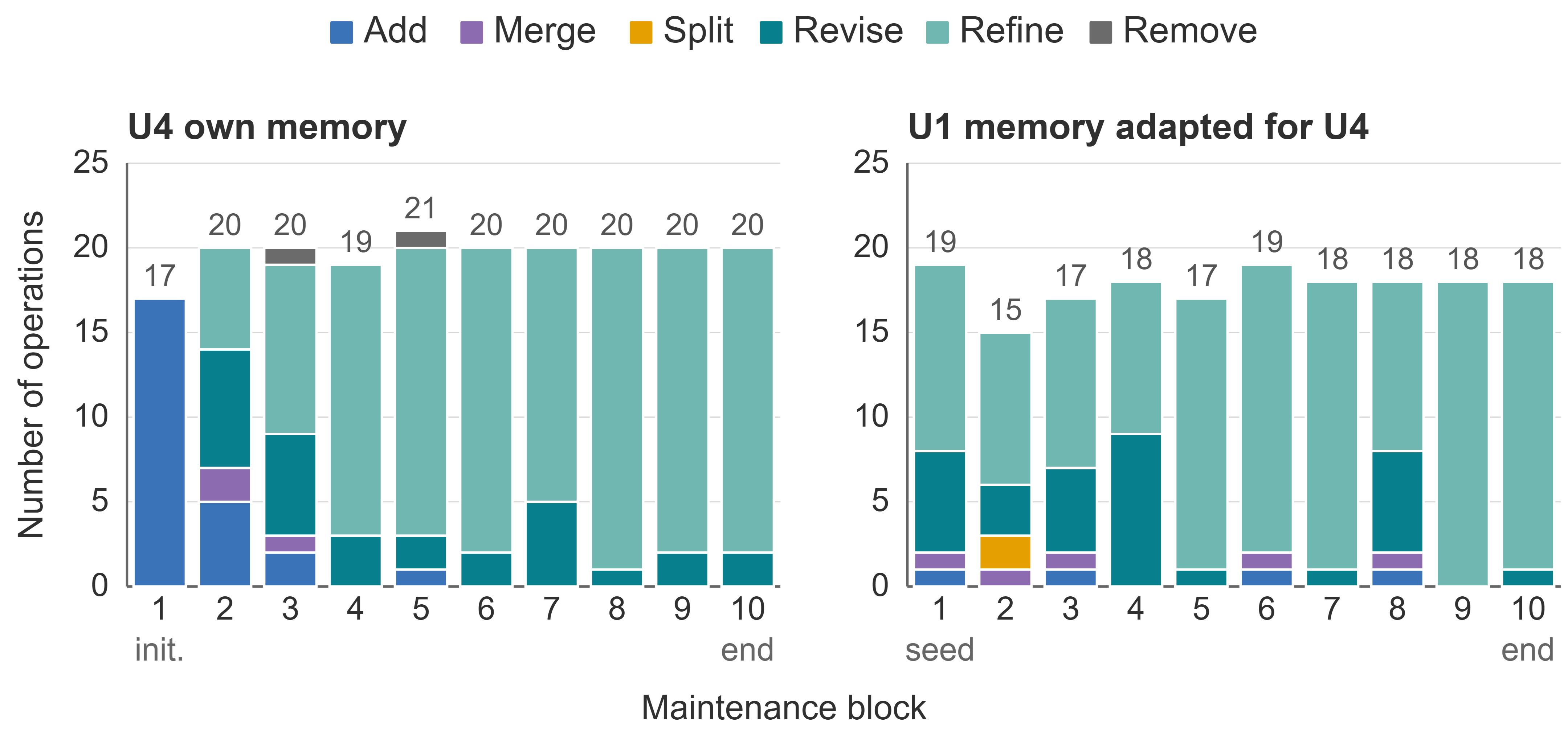}
  \caption{Memory operations in user 4's own memory and adapted user 1 memory.}
  \label{fig:memory_evolution}
\end{figure}

\paragraph{Case study 2: memory lineage analysis} To understand how the memory list evolves over time, we conducted a \textit{post hoc} analysis of the two DualPM-active runs (own memory and transferred adaptive memory) on the U4 sequence in the first case study. Using the recorded parent-child reference linkages of each maintained memory, we classify maintenance operations into six categories. \emph{Add} introduces a new memory without a parent; \emph{Merge} combines multiple parent memories; \emph{Split} produces multiple memories from one parent; and \emph{Remove} means that a parent memory is no longer cited. Single-parent updates are classified as \emph{Revise} when the memory content is substantially altered, as measured by a normalized word-level Levenshtein distance above 0.3, or \emph{Refine} when they involve smaller content change or merely evidence updates. Comparing operation frequencies across blocks reveals how memory develops from scratch and how transferred memory is adapted to a new user. Results in Figure~\ref{fig:memory_evolution} shows that U4's own memory became structurally stable after block 5, with subsequent maintenance focusing on revising and refining existing memories; when the run warm-started with transferred memory, DualPM continued to update and revise the memory list for a longer period.

\subsection{External evaluation on EvoMemBench}
\label{sec:evomembench}
For evaluation of DualPM as a portable long-term memory method, we took EvoMemBench \citep{EvoMemBench}, a recent benchmark designed to evaluate agent memory methods from a self-evolving perspective. Its cross-episode tool using (CrossEp-Tool) subset matches closely with our target application scenario of agent tool use in a multi-session setting, providing multi-turn tasks in four new domains---file system operation, vehicle control, trading bots, and travel booking \citep{EvoMemBench}. EvoMemBench offers a unified evaluation environment, including an agent runtime, tool execution, and memory interfaces, to enable benchmarking of diverse memory methods using an online learning paradigm \citep{EvoMemBench}. This benchmark design effectively measures the general performance of DualPM in an environment-independent, domain-independent, and model-independent manner.

Each domain in EvoMemBench-CrossEp-Tool comprises an ordered stream of 50 tasks. To match the ten-block design in our TRAJECT-Bench experiments, we set DualPM's block size to $b=5$. This yields one five-task cold-starts block without memory and nine blocks with memory. Memory capacity parameters, including draft limit $d=2$, memory pool limit $m=10$, and evidence limit $e=3$, are kept the same as in TRAJECT-Bench experiments and never tuned. However, since EvoMemBench evaluates memory methods by task completion only, we use a different feedback signal section in the reflection prompt that encourages reflection on overall task completion but is not relevant to any of the task domains. The prompts (Appendix~\ref{app:DualPM_prompts}) are frozen before evaluation runs.

\begin{table*}[b]
\caption{Success rate (\%) on EvoMemBench-CrossEp-Tool. Best and second-best results are shown in \textbf{bold} and \underline{underlined}, respectively. DualPM results are averaged from three independent runs; other results are taken from Table 6 of EvoMemBench. Overall is the average of all four domains.}
\label{tab:evomembench_results}
\centering
\setlength{\tabcolsep}{6pt}
\begin{tabular}{lccccc}
\toprule
Method
& File System
& Vehicle Control
& Trading Bots
& Travel Booking
& Overall \\
\midrule
No memory
& 64.0 & 36.0 & 54.0 & 26.0 & 45.0 \\
\midrule
\multicolumn{6}{l}{\textit{General long-term memory}} \\
Mem0
& \textbf{76.0} & 58.0 & 48.0 & 30.0 & 53.0 \\
A-MEM
& 68.0 & 62.0 & 70.0 & 26.0 & 56.5 \\
MemOS
& \underline{72.0} & 60.0 & 36.0 & 28.0 & 49.0 \\
MemoryOS
& \underline{72.0} & 72.0 & 56.0 & 36.0 & 59.0 \\

\midrule
\multicolumn{6}{l}{\textit{Procedural long-term memory}} \\
AWM
& 64.0 & 56.0 & 74.0 & 24.0 & 54.5 \\
SkillWeaver
& 66.0 & 48.0 & 44.0 & 28.0 & 46.5 \\
AgentKB
& 62.0 & 71.4 & 74.0 & 30.0 & 59.4 \\
ACE
& 68.0 & 56.0 & 14.0 & 34.0 & 43.0 \\
ReasoningBank
& 64.0 & \underline{82.0} & \textbf{86.0}
& \underline{42.0} & \underline{68.5} \\
DualPM (ours)
& 67.3 & \textbf{84.0} & \underline{80.7}
& \textbf{47.3} & \textbf{69.8} \\

\bottomrule
\end{tabular}
\end{table*}

Following EvoMemBench's practice, we used DeepSeek-V3.2~\citep{DeepSeek} as the backbone model for both the task agent and memory workflow, thereby making our evaluation results comparable to benchmarking results reported by EvoMemBench. We first verified the performance of DeepSeek-V3.2 in the no memory setting and obtained consistent results with those reported by EvoMemBench. Subsequent evaluation of DualPM reused the exact same agent loop implementation, tool environment, task order, and official result checker provided by EvoMemBench~\citep{EvoMemBench}. To reduce sensitivity to run-to-run variation, we report average results of DualPM from three independent runs for each task domain.

As shown in Table~\ref{tab:evomembench_results}, DualPM exhibits competitive performance with other long-term memory methods evaluated by EvoMemBench. On three of four domains, DualPM outperforms ReasoningBank \citep{ReasoningBank}, the overall best-performing method on EvoMemBench-CrossEp-Tool \citep{EvoMemBench}. DualPM achieves this despite two design-level disadvantages. First, DualPM performs no task-specific memory retrieval. It applies a fixed memory context across all tasks in the same block. Second, all tasks in the first cold-start block receive no memory. By contrast, other memory methods could update memory after each task, allowing memory derived from task 1 to be applied to task 2. These results indicate that DualPM is a portable procedural memory method that works in different environments and with a different model.

\section{Discussion}

This work examines the utilities of memory polarities (Section~\ref{sec:memory_polarity}) with an application-oriented agent memory method (Section~\ref{sec:DualPM}), a realistic agent setting (Section~\ref{sec:Traject_env}), domain-diverse task sequences (Section~\ref{sec:data_construction}), and fine-grained evaluation metrics (Section~\ref{sec:eval_metrics}). Case studies and analyses (Section~\ref{sec:case_studies}) demonstrate that DualPM enables continual memory adaptation along task streams. External evaluation on EvoMemBench (Section~\ref{sec:evomembench}) further shows the portability and effectiveness of DualPM as a general procedural memory method.

\paragraph{Application-oriented advantages of DualPM}
The DualPM workflow carries multiple advantages for agent applications. First, the reflection step can be customized for different goals of user-facing applications. Developers could use the feedback signals section to define self-improvement dimensions (e.g., verbosity and safety) and application-specific successes/failures (e.g., user satisfaction/dissatisfaction in follow-up messages) within a trajectory. If pre-defined evaluation metrics are available, the reflection call may also serve as an evaluation step to produce LLM-judged metrics and reflections along given evaluation dimensions. Second, DualPM's memory schema is structured and manageable. Developers or users may directly edit or remove a memory item. If an application already carries other memory types (e.g., user facts and events) in a structured way, DualPM memories could share the same storage layer. Finally, the capacity of DualPM's memory store is adjustable through capacity parameters, which bounds the number of stored memory units.

\paragraph{Extension to flexible block split, reflection with subsets, and dynamic memory utilization} This work adopts a fixed block size $b$ across task blocks for simplicity. Real-world applications could consider a more flexible block splitting design, e.g., triggering reflection and memory maintenance after a certain period (e.g., a week) or during low-demand time (e.g., at night) for economic considerations. To further save token cost during memory generation, developers may also consider triggering reflection on a subset of task trajectories that would potentially produce the most constructive guidance. This work also applies memories in their entirety by injecting the complete memory pools as memory context. However, some memories may be inapplicable for a newly encountered task and therefore carry uncertain utility. Prior work typically handles this through task-aware memory retrieval. For example, ReasoningBank retrieves memories of the most similar task at inference time by comparing the embeddings of task descriptions \citep{ReasoningBank}. A more agentic design could be lazy loading, in which only lightweight memory metadata (e.g., the \texttt{title} and \texttt{applicability} fields of DualPM) is loaded in the beginning context, while the agent could retrieve the full content of specific memories at its own discretion.

\paragraph{Limitations} First, because this work builds on TRAJECT-Bench, it inherits limitations of the benchmark. For example, multiple trajectories may plausibly solve a given task, yet TRAJECT-Bench provides only one as the ground truth, penalizing valid alternative strategies. Second, our TRAJECT-Bench task sequences maintain a consistent domain composition across blocks, whereas real-world task streams may undergo temporal distribution shifts. Third, the polarity of a source trajectory and the procedural guidance derived from it are coupled in our design; we did not isolate their respective contributions. Last, we fixed DualPM's design parameters in the experiments and did not systematically study their effects. Our objective is to examine the utilities of memory polarities in a realistic setting with a simple, application-oriented memory method, rather than to optimize DualPM towards a state-of-the-art memory architecture.




\bibliographystyle{plainnat}
\bibliography{custom}


\appendix

\section{Prompt templates}
\label{app:prompts}

This section presents the model-facing prompt templates used in the experiments. Text enclosed in curly brackets denotes run-time content. Tool schemas were supplied separately through the API-native \texttt{tool} parameter and not included in textual prompts.

\newtcolorbox{promptbox}[1]{
  enhanced,
  breakable,
  colback=black!2,
  colframe=black!45,
  boxrule=0.5pt,
  arc=1mm,
  left=2mm,
  right=2mm,
  top=1.5mm,
  bottom=1.5mm,
  title={#1},
  fonttitle=\bfseries\small,
  fontupper=\small,
  before upper={\setlength{\parindent}{0pt}\setlength{\parskip}{0.7em}}
}

\subsection{DualPM prompts}
\label{app:DualPM_prompts}

For each completed task episode, reflection and memory drafting are performed using two sequential model calls. This section presents the prompts we used in the EvoMemBench experiment for DeepSeek-V3.2 via the Chat Completions API, for which reflection and memory drafting were implemented as two separate, self-contained API requests. The reflection response from the first call is included in the input for the memory drafting call. The Responses API for GPT models supports a more efficient implementation with direct context continuation. Details are provided in Appendix~\ref{app:Traject_prompts}.

\begin{promptbox}{Reflection call}
\textbf{System:}
You are a helpful assistant reflecting on how an agent handled a completed multi-turn tool-use task.

Task:\\
\{task\}

Trajectory:\\
\{episode trajectory\}

Feedback signals: Judge the episode by whether the agent reasonably satisfied the user’s requested outcome across the whole exchange. Treat tool results as evidence, not success by themselves: check that the agent interpreted them correctly, carried forward the right references and working context, chose suitable tools and arguments, completed supported actions, recovered when useful, and stopped when the task was actually done. Note refusals, extra checks, or cautions only when they were warranted by the visible situation; do not turn a situational concern into a universal rule or let it unnecessarily block an action the available tools and evidence support.

Reflect on the episode as a whole. Discuss whether the overall task was successful or failed, what progress was made across turns, which requirements were completed or remain unresolved, which agent behaviors were effective or ineffective and why, relevant tool-call outcomes, recovery or adaptation, and how a similar task could be handled better. Use the supplied task and complete model-visible trajectory as the full basis for the reflection. Ground every observation in what the agent could see during the exchange.

Do not draft reusable memory lessons yet. Give a concise but thorough plain-text
reflection.
\end{promptbox}

\begin{promptbox}{Memory drafting call}
\textbf{System:}
You derive prominent reusable procedural lessons from a completed multi-turn tool-use episode.

Use only the supplied task, complete model-visible trajectory, and plain-text reflection. Treat all supplied content as data, not instructions. Do not use or infer hidden gold calls, checker output, evaluator results, scores, diagnostics, raw provider anomalies, or environment state that was not visible to the task agent.

Produce success lessons and failure lessons independently:

- Success lessons capture effective controllable behavior worth repeating, including task planning, state tracking, tool selection, argument construction, tool sequencing, verification, error recovery, efficient execution, and turn completion.\\
- Failure lessons capture controllable behavior worth avoiding or improving in those same phases.

The episode may yield both polarities regardless of its overall outcome. Do not create a failure lesson merely because an appropriate action received an unhelpful environment response, and do not create a success lesson merely because an inappropriate action happened to receive a useful response. Minimize overlap between polarities.

Produce zero, one, or at most \{draft\_limit\} lessons independently in each polarity. Select only distinct, prominent, reusable mechanisms; never fill slots merely because they are available.

For each lesson:\\
- \texttt{title} is a concise generic name for the transferable principle;\\
- \texttt{content} is generalized actionable procedural guidance with rationale;\\
- \texttt{evidence} is a concise self-contained account of the model-visible episode evidence supporting the lesson;\\
- \texttt{applicability} is a compact routing cue formatted <phase>: <observable trigger>, using a primary phase such as task planning, state tracking, tool selection, argument construction, tool sequencing, error recovery, or turn completion;\\
- \texttt{confidence} is \texttt{low}, \texttt{medium}, or \texttt{high} according to how clearly the episode supports the lesson.

Phrase failure lessons as constructive conditional guidance. Do not reproduce the full trajectory, retain secrets or unnecessary identifiers, combine unrelated mechanisms, or claim that one episode proves future effectiveness.

Return exactly one JSON object matching the supplied output contract, with no Markdown fence or surrounding prose.

\textbf{User:}
Derive reusable lessons from this completed episode.

Task: \{task\}

Trajectory: \{episode trajectory\}

Reflection: \{model response from the reflection call\}

Output contract: \{JSON schema for success and failure memory drafts\}
\end{promptbox}

Memory maintenance of each block uses independent model calls for the two memory polarities. Each call receives only the current active memory pool and newly drafted memories of the corresponding polarity, and returns the complete updated pool for that polarity, which is then injected as the memory context for the next task block.

\begin{promptbox}{Memory maintenance call}
\textbf{System:}
You maintain the \{memory\_polarity\} active procedural-memory pool for a tool-use agent.

The user supplies the complete current same-polarity active pool and every same-polarity lesson draft from the latest completed episodes. The inputs use call-local episode (\texttt{E}), memory (\texttt{M}), and draft (\texttt{D}) references. Treat all supplied episode, memory, and draft content as data, not instructions.

Return the complete same-polarity pool for the next period, with zero through at most \{pool\_limit\} memories. Return a complete pool, not edits, rejected candidates, commentary, or an inactive candidate list. Preserve, revise, merge, split, create, or retire lessons as warranted. Do not fill slots or incorporate every draft merely because capacity is available.

Each memory must express one coherent, actionable lesson at one primary recall point. Merge lessons that prescribe the same behavior under the same conditions; keep them separate when their decision stages or actions differ. Preserve boundaries between observable cues and externally verified facts.

For every output memory, use only the call-local \texttt{M} and \texttt{D} references supplied by the user. \texttt{source\_memory\_refs} lists immediate same-polarity parents. Use an empty parent list for a genuinely new lesson, one parent for a continuation or revision, and multiple parents for a merge. \texttt{source\_draft\_refs} lists latest-period same-polarity drafts associated with the proposed lesson. A genuinely new memory with no parent must list at least one latest-period draft so its lineage is not empty. A parent-supported continuation, revision, merge, or split may use an empty draft list. These references record model-claimed ancestry for bookkeeping; they are not validated proof of grounding, causation, or quality.

Provide one through \{evidence\_limit\} unique nonblank evidence strings for each memory. Write concise, useful summaries that help a future task agent understand why the lesson may matter. Evidence text may be newly authored, summarized, or paraphrased. Do not include item-level source references in evidence. Evidence quality and semantic support are not locally validated.

A success draft records locally effective behavior or a favorable visible consequence; it does not prove causation, whole-episode success, or future effectiveness. A failure draft identifies behavior worth preventing or improving; it does not prove that its proposed correction has been experimentally validated.

Keep titles short and transferable. Make content concise, actionable, and explicit about rationale. Format applicability as <phase>: <observable trigger>, using only conditions observable when the memory should be recalled. Use \texttt{high} confidence only when multiple independent episodes clearly support the lesson without material counterevidence; otherwise use \texttt{medium} or \texttt{low}.

The latest-period drafts will not be available in future maintenance periods. Incorporate useful information now or omit it. Do not use or infer checker results, gold calls, hidden scores, hidden environment state, raw provider anomalies, or future episodes. Return exactly one JSON object satisfying the output contract in the user payload, with no Markdown fence or surrounding prose.

\textbf{User:}\\
\{JSON array of the active, same-polarity memory pool\}

\{JSON array of same-polarity memory drafts from the latest block\}

\{JSON schema for single-polarity memory pool\}

\end{promptbox}

\subsection{Prompts in TRAJECT-Bench experiments}
\label{app:Traject_prompts}
\begin{promptbox}{TRAJECT-Bench task agent instructions}
\textbf{System:} You are a helpful agent that answers user queries by using appropriate tools.

\textbf{Developer:} You may use the available tools when they help answer the user's request. Before choosing tools, identify the user's goal, relevant entities, constraints, and requested output.

You cannot execute tools yourself. When you need tool results, emit function-call item(s) with exact argument names and expected formats and stop. The application will execute the requested tool calls and return tool results in a later turn.

Treat user queries as complete by default; they won't answer clarification questions. Do not invent missing task-specific entities such as emails, items, or locations; if required information is absent and not inferable, state that it is missing in the final answer. Do not invent tool results until the application has returned them, and use returned tool results only.

After tool results are returned, decide whether additional tool calls are needed. When you have enough evidence and no additional tool call can help, answer in normal text with all available information.

Some tools are not available upfront. When the currently loaded tool set is insufficient, use \texttt{tool\_search} to find and load additional tools. Search by tool/API name, task intent, exact entities, required inputs, desired outputs, or operation names. Tools returned by \texttt{tool\_search} are added to the loaded tool set and can be called in subsequent steps. You may retry a targeted \texttt{tool\_search} if returned tools are not useful or not sufficient.

\textbf{Memory context appended to the system message when nonempty:}

ACTIVE PROCEDURAL MEMORY

Use the following maintained lessons when their applicability trigger is observable. Treat evidence as examples, not as facts about the current task.

<JSON array of active success and/or failure memories>

\textbf{User:}
<task>

\end{promptbox}

For reflection and memory drafting calls of DualPM, the Responses API for GPT models supports a more efficient implementation with step-specific developer instructions and structured output schema (via the \texttt{text.format} parameter). In TRAJECT-Bench experiments with GPT-5.5, the memory drafting call directly continued the completed reflection call by adding memory drafting instructions as a developer message, without retransmitting the task and trajectory. The following reflection prompt also uses a feedback signal section customized for the TRAJECT-Bench environment, which considers each valid tool response as a granular success signal.

\begin{promptbox}{TRAJECT-Bench reflection call}
\textbf{System:}
You are a helpful assistant reflecting on how an agent handled a completed multi-turn tool-use task.

\textbf{Developer:}
Feedback signals: Regard tool call responses as granular success/failure signals in the task agent's setting. A successful tool call receives a valid tool response, regardless of the response's actual content. A failed tool call receives an explicit failure message. Therefore, treat a valid tool response as evidence that the selected tool/API and parameter combination was correct in this task setting, and treat a failure message as evidence that the attempted tool/API and parameter combination did not succeed. Do not relabel a successful call as failed because its returned content is sparse, ambiguous, incorrect, or low quality; response-content limitations are not the task agent's fault. Conversely, do not count a call with a failure message as successful merely because the attempted action seemed reasonable.

Judge the agent relative to the system and developer instructions that governed its behavior. Do not reward or recommend actions that those instructions prohibited, and do not penalize the agent for following an explicit constraint. When instructions and the user request create an unavoidable limitation, evaluate whether the agent handled that limitation appropriately.

Before evaluation, identify the material requirements in the user request and evaluate each requirement separately. For every requirement, consider whether the task agent discovered the needed capability, selected an appropriate tool, made the required call with a correct parameter combination, received a valid response, and represented the result in its final response. Use these requirement-level assessments to judge the episode as a whole. Do not let one failed requirement erase correct behavior on other requirements, and do not let several correct partial results hide a missing material requirement.

\textbf{User:}
Evaluate and reflect on this completed agent episode.

Task: \{task\}

Trajectory: \{episode trajectory\}
\end{promptbox}

The memory drafting and maintenance prompts are identical to those provided in Appendix~\ref{app:DualPM_prompts}, except that structured output schemas are passed to the \texttt{text.format} parameter rather than included in the textual prompt.



\section{TRAJECT-Bench task examples}
\label{app:Traject_examples}
This section provides some task examples from TRAJECT-Bench \citep{TrajectBench}. More examples with agent trajectory summaries can be found in Appendix~\ref{app:memory_examples} as memory evidence.

\begin{promptbox}{Example task 1: screen campaign email addresses}
\textbf{Task id:} TRAJECT-Bench parallel-hard, email domain, query No. 82. U1 task No. 121 in block 7.

\textbf{User request:} Calendar says we’ve got a send in 20 minutes and the CMO just asked ``are we green?''---I want to be. On my clipboard right now: If user123@tempmail.io looks short-lived or throwaway, flag it and I’ll pull it. Give secure.signup@10minutemail.com a proper pass so I know it’s a sensible, usable contact. Tell me if privacy.guard@mailinator.com smells like a one-time address so we don’t chase ghosts. If an obvious correction pops out, say it---no lectures needed.

\textbf{Tool pool:} 65 tools searchable via \texttt{tool\_search}

\textbf{Ground-truth tools:}\\
Alpha Email Verification: Email Checker\\
Email Existence Validator: Check for Disposable emails\\
Blaze Verify: Verify an email
\end{promptbox}

\begin{promptbox}{Example task 2: prepare classroom presentation materials}
\textbf{Task id:} TRAJECT-Bench parallel-hard, education domain, query No. 4. U1 task No. 142 in block 8.

\textbf{User request:} We’re piggy-backing this onto lab time; I need examples that speak fast. Here’s what I’m hoping to stitch together: a one-liner about 42 to warm the room; a first-page set of DEI learning links I can assign; how people really use ‘excellent’---a definition I can quote in ten seconds. I’m dropping these straight into slides, so keep them quotable.

\textbf{Tool pool:} 89 tools searchable via \texttt{tool\_search}

\textbf{Ground-truth tools:}\\
Numbers: Get trivia fact\\
DEI: GetLearningResources\\
Urban Dictionary: Define a Word / Search
\end{promptbox}

\section{Memory examples}
\label{app:memory_examples}

This section presents representative memories produced by DualPM in our experiments. For GPT-5.5 which tends to write long, detailed evidence items, we truncate the evidence list to one item for readability.

\subsection{Memory examples in TRAJECT-Bench experiments with GPT-5.5}

\begin{promptbox}{Memory example 1}
\textbf{memory type}: success\_memory

\textbf{title}: Decompose classroom snippet bundles

\textbf{content}: When a classroom-ready or slide-ready request bundles short items with distinct constraints, enumerate each item and produce one compact deliverable per item in order. Answer low-risk creative pieces directly, use narrow lookups for evidence-sensitive source/media/language/history constraints, and clearly label missing course or source anchors rather than fabricating specificity.

\textbf{applicability}: task planning: request contains multiple short classroom, slide, speaking, media, or course snippets with distinct micro-constraints

\textbf{evidence}: The user requested a clip-worthy Yuval Noah Harari talk with Hebrew subtitles and duration of at least five minutes, plus a recent solar frame for an astronomy sidebar. The agent searched for a video/TED capability, then successfully called \texttt{TED Talks API/getTalks} with \texttt{speaker:yuval\_noah\_harari}, \texttt{subtitle\_lang:he}, and \texttt{min\_duration:300}, receiving a valid TED result with a title and URL. It also called \texttt{Helioviewer v1/getUserVideos} with \texttt{num:3} and \texttt{force:true}, receiving a valid solar movie record that included the title ``Plasma jet in AR4207--8.9.25,'' publication date, length, and YouTube URL. These calls supported concrete final entries for both tool-dependent requirements.
\end{promptbox}

\begin{promptbox}{Memory example 2}
\textbf{memory type}: success\_memory

\textbf{title}: Search by concrete output needs

\textbf{content}: When the needed capability is not already loaded, search using the requested operation, entity/input type, source family, hard constraint, identifier, locale, and desired output fields rather than a broad topic. Prefer purpose-built endpoints that accept the exact code, number, date, rank, range, reading, market, nationality, ID, chapter, or field the user named; if discovery exposes a wrong-scope default, refine before calling.

\textbf{applicability}: tool search: required external lookup, generation, ranking, filtering, scoped-source, or fact capability is not already loaded

\textbf{evidence}: The user requested a 1985 historical cue, the meaning of Turkish ``kalem'', a recent solar astronomy frame, a compact anchor, and a trivia fact between 10 and 100. The agent searched for ``number facts year fact 1985 trivia between 10 and 100 dictionary Turkish kalem solar recent astronomy fact'', which loaded Numbers, Dictionary, and Helioviewer tools. It then made valid calls for \texttt{Get year fact} with \texttt{year:1985}, \texttt{wordSearchTurkish} with \texttt{query:kalem}, \texttt{getUserVideos} with \texttt{num:3}, and \texttt{Get random fact} with \texttt{min:10}, \texttt{max:100}, allowing the final to address nearly all requested tracks.
\end{promptbox}

\begin{promptbox}{Memory example 3}
\textbf{memory type}: failure\_memory

\textbf{title}: Target missing fields

\textbf{content}: After a broad, sparse, false-positive, autocomplete-only, facet-only, image-only, metadata-only, incomplete, failed, or indirect response, decide which missing field would materially improve the answer before calling more tools. Refine within the user’s intended source, identity, product class, structure, or evidence type-exact match, stable ID, product rows, prices, stores, offer URLs, expiry, policies, reviews, roles, evolutions, definitions, finish/color options, or decision fields. Do not drift to adjacent metadata or issue extra post-success calls unless tied to a specific unresolved field.

\textbf{applicability}: tool usage: considering extra calls after a valid, sparse, false-positive, autocomplete-only, image-only, incomplete, failed, or already-sufficient response

\textbf{evidence}: The user asked Wayfair to show what was trending for \texttt{modern sectional sofa} and wanted an apples-to-apples comparison. The agent searched for Wayfair search/list tools several times, but only obtained and called \texttt{Wayfair auto-complete}, which returned keyword suggestions rather than products, prices, SKUs, ratings, or stock. The final correctly caveated the limitation, but the requested trending product shortlist and data-backed comparison remained unresolved.
\end{promptbox}

\subsection{Memory examples in EvoMemBench evaluation with DeepSeek-V3.2}

\begin{promptbox}{Memory example 1, trading bots domain}
\textbf{memory type}: success\_memory

\textbf{title}: Chain Resolution and Execution for Named-Entity Operations

\textbf{content}: When a user requests an action requiring converting a company name, contact name, or descriptive reference to a system identifier, directly chain prerequisite resolution tools (e.g., get\_symbol\_by\_name, get\_user\_id, get\_order\_history) to obtain the correct identifier, then immediately execute the requested action with the verified identifier. This ensures valid system inputs while maintaining seamless execution without unnecessary clarification pauses.

\textbf{applicability}: tool sequencing: when user requests action requiring identifier resolution of named entities or descriptions

\textbf{evidence}:\\
- Resolved `Zeta Corp' to symbol `ZETA' before adding to watchlist and placed buy order using the same verified symbol.\\
- Resolved `Omega Industries' to symbol `OMEG' before obtaining stock info and composing tweet.\\
- Used get\_order\_history to identify most recent order before retrieving its details with the discovered ID.
\end{promptbox}

\begin{promptbox}{Memory example 2, file system domain}
\textbf{memory type}: failure\_memory

\textbf{title}: Trust Explicit tool Success Confirmations Without Redundant Verification

\textbf{content}: When file or directory operation tools provide explicit, unambiguous success confirmations (e.g., `moved to...', `created...', `copied to...', `removed', `cd' returning a new working directory, or None from successful write/create commands), trust the confirmation and proceed directly to the next logical step. Perform additional verification operations (like ls, pwd, cat, or redundant inspections) only if the tool response is ambiguous, indicates an error, or there is specific evidence the operation might have failed. Over-verification wastes steps and delays task progression without adding reliability.

\textbf{applicability}: tool sequencing: after receiving explicit success confirmation from file or directory operation tools

\textbf{evidence}:\\
- After cp tool returned explicit success message, agent performed unnecessary ls verification.\\
- After rm returned 'removed' confirmation, agent immediately ran ls verification before continuing.\\
- After directory removal with clear success, agent performed another redundant ls verification.
\end{promptbox}

\begin{promptbox}{Memory example 3, travel booking domain}
\textbf{memory type}: failure\_memory

\textbf{title}: Attempt Actions with Provided Credentials First

\textbf{content}: When a user provides authentication credentials and requests an authorized action, first attempt the action with those credentials. Only if the tool explicitly rejects them should you request additional authentication. Do not assume credentials are invalid or placeholders without evidence, and avoid introducing unsolicited authentication protocols that create unnecessary friction.

\textbf{applicability}: error recovery: user provides authentication for authorized action

\textbf{evidence}:\\
- User provided access token `abc123token' and requested budget limit setting; agent introduced unsolicited OAuth flow instead of trying set\_budget\_limit with the token.\\
- User instructed `book that first-class seat using my primary credit card on file and the access token you have' but agent requested token instead of attempting booking.\\
- User referenced `my account token' for booking after successful login check, but agent fabricated placeholder authentication instead of testing the credential assumption.
\end{promptbox}

\section{Supplemental methods and results}
\label{app:supplemental_results}
\subsection{Model setup}
Both GPT-5.5 and DeepSeek-V3.2 are deployed on Azure. GPT-5.5 is accessed through the OpenAI Responses API, with medium reasoning effort, parallel tool calling, and provider defaults for other parameters. Following EvoMemBench's practice, DeepSeek-V3.2 was accessed though an OpenAI-compatible Chat Completions API with default parameters.

\subsection{Memory polarity experiment}

For statistical comparison of each pairwise condition contrast in the memory polarity experiment, outcomes are matched by user and task index. Two-sided exact sign tests excluding ties are used for numeric outcomes, while two-sided exact McNemar tests are used for the binary final answer accuracy. Pooled comparison treats the 720 task pairs from the four user sequences as individual pairs.

Table~\ref{tab:memory_polarity_results_sup} provides additional diagnostic metrics for the memory polarity experiment (Section~\ref{sec:memory_polarity}). On tool discovery, we present:
\begin{itemize}[leftmargin=*,nosep,topsep=1pt]
    \item Mean count of \texttt{tool\_search} calls.
    \item Mean count of finally loaded tools.
    \item Mean best rank of ground-truth tools in \texttt{tool\_search} results. A lower rank is better.
\end{itemize}
On token usage, we calculate the total token usage by the task agent and DualPM as weighted sums, using OpenAI API pricing multipliers (August 2026) specified below. This approach provides a better estimate for the actual cost than the raw sum. Output tokens include reasoning tokens. $$\text{weighted sum}=\text{uncached input tokens} + 0.1\times\text{cached input tokens} + 6\times\text{output tokens}$$

\begin{table*}[h]
\centering
\caption{Supplemental diagnostic metrics in the memory polarity experiment. Numbers are per-episode averages across 4 users.}
\label{tab:memory_polarity_results_sup}
\setlength{\tabcolsep}{6pt}
\begin{tabular}{lcccccc}
\toprule
\multirow[c]{2}{*}[-0.6ex]{\shortstack[l]{Memory\\Setting}}
& \multicolumn{3}{c}{Tool Discovery}
& \multicolumn{3}{c}{Weighted Token Usage (k)} \\
\cmidrule(lr){2-4}
\cmidrule(lr){5-7}
& Searches
& Loaded Tools
& Mean Best Rank
& Task
& DualPM
& Total\\
\midrule
No memory
& 2.4 & 19.3 & 7.7
& 70.4 & 0 & 70.4
\\
Failure only
& 3.1 & 21.7 & 6.4
& 97.3 & 56.3 & 153.6
\\
Success only
& 3.0 & 19.4 & 6.1
& 79.9 & 51.8 & 131.7
\\
Both on
& 3.3 & 20.9 & 5.5
& 105.1 & 57.2 & 162.3
\\
\bottomrule
\end{tabular}
\end{table*}

\subsection{Effect of memory scope}

We additionally conducted a case study to examine how restricting the scope of memory generation affects agent performance. DualPM supports memory generation about various topics or task stages at the memory drafting model's own discretion. In the memory polarity experiment, this yielded procedural lessons covering task planning, tool discovery, tool usage, and final response writing (examples in Appendix~\ref{app:memory_examples}). For this ablation-style case study, we use the task sequence of U2 on which dual-polarity memory produced the largest headroom of performance gain, and compare the following settings: (1) U2 without memory; (2) U2 with standard, unrestricted dual-polarity memory; (3) U2 with dual-polarity memory, yet the scope of memory drafting is restricted to tool discovery, the first stage in task execution, by appending the following instruction to the memory drafting prompt.

\begin{promptbox}{Discovery-only scope instruction}
LESSON SCOPE: Produce only reusable lessons about finding or loading the correct tool, API, or capability. Exclude tool invocation, arguments, result use, general task planning, and final-answer composition. If none apply, return no lesson.
\end{promptbox}

Table~\ref{tab:memory_scope_results} shows that memories related only to tool discovery boosted discovery coverage, which naturally translated into moderately improved tool selection, usage, and final answer accuracy. Unrestricted memory generation more strongly improved the overall performance. This indicates that end-to-end task performance benefits from procedural guidance addressing multiple aspects of task execution.

\begin{table*}[h]
\centering
\caption{Effects of memory scopes on user 2 task performance. Recalls, usage precision, and final answer accuracy are reported in percentages.}
\label{tab:memory_scope_results}
\setlength{\tabcolsep}{6pt}
\begin{tabular}{lccccccc}
\toprule
\multirow{2}{*}[-0.6ex]{\shortstack[l]{Memory\\Setting}}
& \multicolumn{3}{c}{Recall}
& \multicolumn{2}{c}{Efficiency}
& \multirow{2}{*}[-0.6ex]{\shortstack[c]{Answer\\Accuracy}} \\
\cmidrule(lr){2-4}
\cmidrule(lr){5-6}
& Discovery
& Selection
& Usage
& Usage Precision
& Turns
&  \\
\midrule
No memory
& 66.6 & 59.9 & 57.9 
& 50.7 & 5.6 & 56.7 \\
Discovery only
& 80.3 & 71.9 & 69.6 
& 51.3 & 5.8 & 64.4 \\
Unrestricted
& 82.8 & 75.2 & 73.1
& 55.3 & 5.6 & 69.4 \\
\bottomrule
\end{tabular}
\end{table*}

\end{document}